\documentclass[graybox]{svmult}

\usepackage{type1cm}        
\usepackage{makeidx}         
\usepackage{graphicx}        
\usepackage{multicol}        
\usepackage[bottom]{footmisc}

\usepackage{newtxtext}       %
\usepackage{newtxmath}       

\usepackage{times}
\usepackage{latexsym}

\usepackage[T1]{fontenc}

\usepackage[utf8]{inputenc}

\usepackage{microtype}

\usepackage{graphicx,color}

\usepackage{float}
\usepackage{adjustbox}

\newcommand{\BNAME}{{\it FCC}}
\newcommand{\TNAME}{{\bf FCC}}

\makeindex             

\begin{document}

\title*{\TNAME: Fusing Conversation History and Candidate Provenance for Contextual Response Ranking in Dialogue Systems}
\titlerunning{FCC for Contextual Response Ranking} 
\author{Zihao Wang, Eugene Agichtein and Jinho Choi}
\institute{Zihao Wang \at Emory University, 201 Dowman Dr, Atlanta, \email{zihao.wang2@emory.edu}
\and Eugene Agichtein \at Emory University, 201 Dowman Dr, Atlanta, \email{eugene.agichtein@emory.edu}
\and Jinho Choi \at Emory University, 201 Dowman Dr, Atlanta, \email{jinho.choi@emory.edu}}

%
%
\maketitle

\abstract{
Response ranking in dialogues plays a crucial role in retrieval-based conversational systems. 
In a multi-turn dialogue, to capture the gist of a conversation, contextual information serves as essential knowledge to achieve this goal. 
In this paper, we present a flexible neural framework that can integrate contextual information from multiple channels.
Specifically for the current task, our approach is to provide two information channels in parallel, \textbf{F}using \textbf{C}onversation history and domain knowledge extracted from \textbf{C}andidate provenance (\TNAME), where candidate responses are curated, as contextual information to improve the performance of multi-turn dialogue response ranking.
The proposed approach can be generalized as a module to incorporate miscellaneous contextual features for other context-oriented tasks. 
We evaluate our model on the MSDialog dataset widely used for evaluating conversational response ranking tasks. 
Our experimental results show that our framework significantly outperforms the previous state-of-the-art models, improving Recall@1 by 7\% and MAP by 4\%. 
Furthermore, we conduct ablation studies to evaluate the contributions of each information channel, and of the framework components, to the overall ranking performance, providing additional insights and directions for further improvements.
}
\section{Introduction}
\label{sec:introduction-related-work}

\begin{figure}[!htbp]
\centering
  \includegraphics[width=0.9\textwidth]{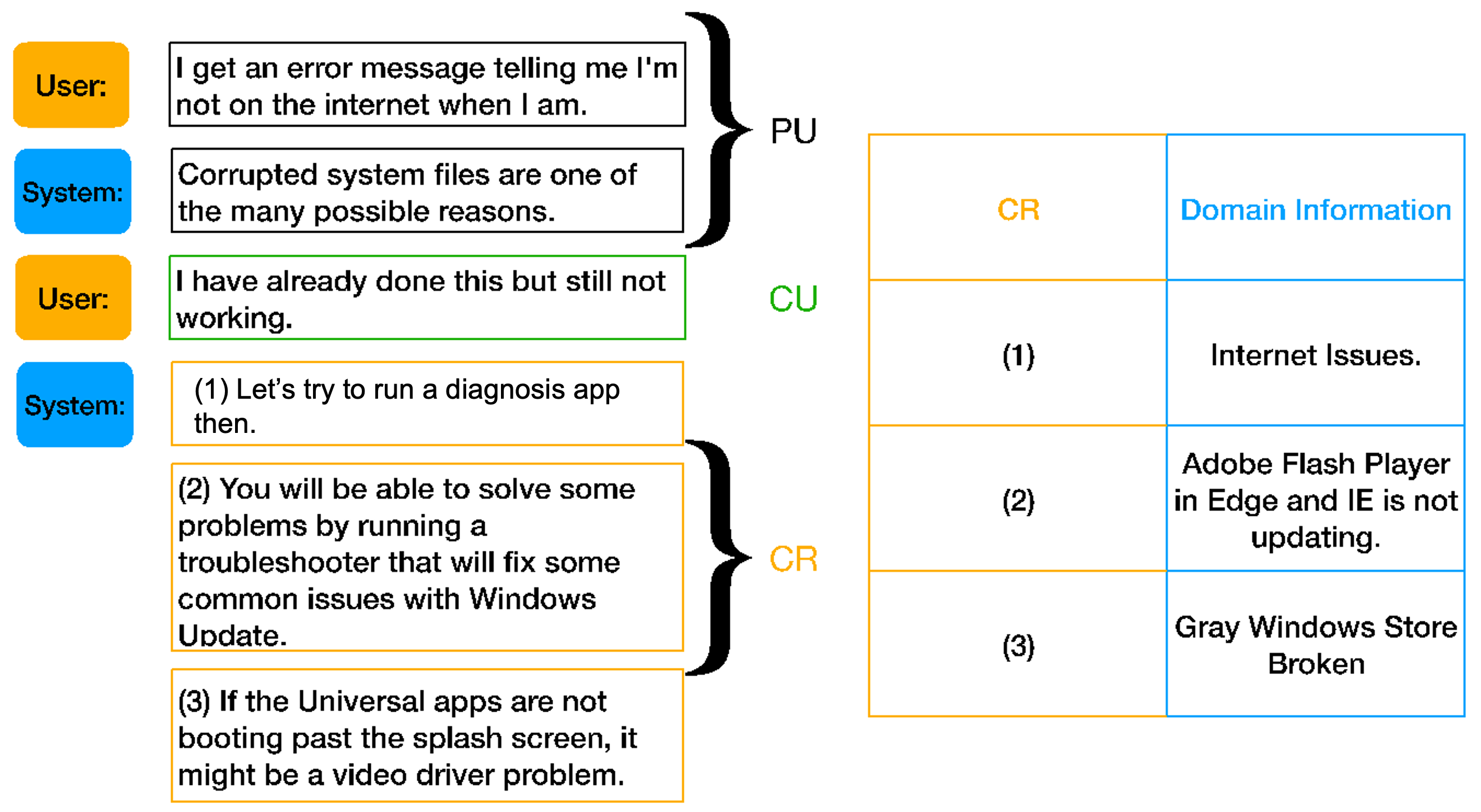}
  \caption{MSDialog conversation example. The abbreviations denote the following. CU: current utterance, PU: previous utterances, CR: candidate responses.}
  \label{fig:dialogue-flow}
\end{figure}

Response ranking is an essential part of dialogue systems \cite{wang2017emersonbot,ahmadvand2018emory}, and plays a critical part in information- or search-oriented dialogues where responses may come from diverse yet usually designated sources.
As shown in Fig. \ref{fig:dialogue-flow}, candidate (1) is the true response, while the other two are negative candidates. 
Semantically, it is reasonably easy to differentiate between candidates (2) and (3). 
However, it is challenging to differentiate candidates (1) and (2), since they describe the same procedure with minor differences. 
In these cases, it is critical to have other surrounding knowledge, such as contextual information, to make distinguishing decisions.

Previous research \cite{zhou2018multi,yang2020iart} has extensively considered modeling of conversation history with external knowledge \cite{yang2018response} for a better understanding of the conversation flow. 
However, most of the previous work did not take into account the {\em source} (provenance) information of candidate responses, such as the domain information where domain-related candidates are curated. This domain knowledge was ignored or treated separately excluded from the overall ranking model, especially in component- and retrieval-based conversation systems \cite{ram2018conversational}.

Instead, we show that it is significant to jointly model the conversation history and domain information from the provenance of candidate responses, as input to built-in parallel information channels in the ranking model, which would allow the model to benefit from both sources of evidence. To validate our idea, implemented as the \BNAME\ models in this paper, we compare experimental results with previous state-of-the-art models, in addition to ablation studies to analyze the effect of domain information of candidates to response ranking performance.
Our experiments are performed on the established benchmark MSDialog dataset, widely used for conversation ranking tasks. 
Our results show that our model significantly improves the performance, by 3-8\% margins on recall@1, recall@2, and MAP, over the previous state-of-the-art models. 
Our ablation studies confirm that domain information of candidates have their advantages over the state-of-the-art models.
Furthermore, we improve the modeling of conversation history by implementing the self-attention mechanism on previous turns and validate its effect to ranking performance by ablation studies.

In this paper, we tackle the response ranking problem by introducing a new aspect, as the candidate provenance, to the end-to-end response ranking pipeline. In addition, we apply the self-attention to model conversation dependency related to previous utterances. Therefore, our contributions can be summarized as: (1) proposing a extensible framework that incorporates domain information associated to candidate response; 
(2) applying self-attention layers to improve the modeling of temporal relationship on conversation history;
(3) conducting studies on the impact of domain information of candidate responses and self-attention layers to the ranking performance.

\section{Related Work}
\label{sec:related work}
We now briefly review related work to place our contribution in context. 
First, we review general learning to rank approaches, which we adapt to the conversational setting. 
Second, we summarize the most recent response ranking models as a transition to our model. 
Next, we review topic modeling and classification in dialogues as it is important and relevant to response ranking in dialogue systems.
Last, we review ranking tasks integrating external knowledge.

\subsection{Learning to rank} 
\label{ssec:learning to rank}
Learning to rank approaches have applications in various fields, such as information retrieval and natural language processing. 
BM25 \cite{robertson1995okapi} and its variants have been widely received as reliable baseline methods. 
Later, supervised machine learning was adapted to ranking tasks, such as the SVM ranking model proposed by \cite{shashua2003ranking}. 
As neural networks started arising, Ranknet\cite{burges2005learning} and LambdaMart\cite{wu2010adapting} were a series of improvements based on gradient descent methods. 
However, these algorithms highly rely on the richness of extracted features, while feature selection methods often compromise semantic meanings. 

\subsection{Neural response ranking models}
\label{ssec:neural response ranking models}
The upsurge of Word2Vec\cite{mikolov2013distributed} and the development of neural network models facilitated learning-to-rank performance, and they are quickly adapted to dialogue response ranking tasks. 
Variations of Convolutional Neural Networks (CNN) \cite{yan2016learning},  Recurrent Neural Networks (RNN) \cite{luan2016lstm}, and the combination of the two \cite{chung2014empirical} have been explored to push the frontline forward. 
Most recently, the sequential matching network (SMN) \cite{wu2016ranking}, deep matching network with external knowledge \cite{yang2018response}, deep attention model \cite{Qu_2018} and the intent-aware model \cite{yang2020iart} have achieved state of the art respectively. 

\subsection{Topic modeling and classification in dialogues} 
\label{ssec:topic modeling and classification dialogues}
Topic modeling and classification are critically important to understand users' topics of interest in a conversation, and it is critical to a dialogue system to acquire candidates from knowledge sources based on topic modeling and classification.
\cite{jokinen1998context} defined topic trees to use topical information for conversational robustness.
Latent Dirichlet Allocation (LDA) was applied by \cite{yeh2014topic} to detect topics in conversational systems.
However, when applied to dialogues, unsupervised models can only infer topics from lexical statistics, which are not always consistent with conversation context.
Supervised methods such as the supervised LDA by \cite{NIPS2007_3328} and a Deep Average Network-based model \cite{guo2018topicbased} further improved topic understanding in either text or dialogues.
Most recently, \cite{ahmadvand2019concet} proposed an entity-aware topic classification model to facilitate the understanding of topics with entities.
After all, the above models missed the link between topics and conversation context.

\subsection{Ranking with integration of external knowledge}
\label{ssec:ranking_with_integration_of_external_knowledge}
Integration of external knowledge has a long history in document ranking and retrieval tasks \cite{berger2017information, cormack2011efficient, dalton2014entity}. 
Various sources of knowledge are utilized to improve the performance of ranking. \cite{hovy2001use} uses well-constructed WordNet and QA typology to improve performance on a Question-Answerign system. 
Wikipedia was used as an external knowledge to improve the document clustering tasks by \cite{hu2009exploiting}. 
\cite{xiong2017jointsem} incorporates entities for document ranking.

In this paper, we utilize both knowledge source information associated to candidate responses, and conversation history to perform the response ranking task in a multi-turn conversation.
\begin{figure*}[!htbp]
\centering
  \includegraphics[width=0.9\textwidth]{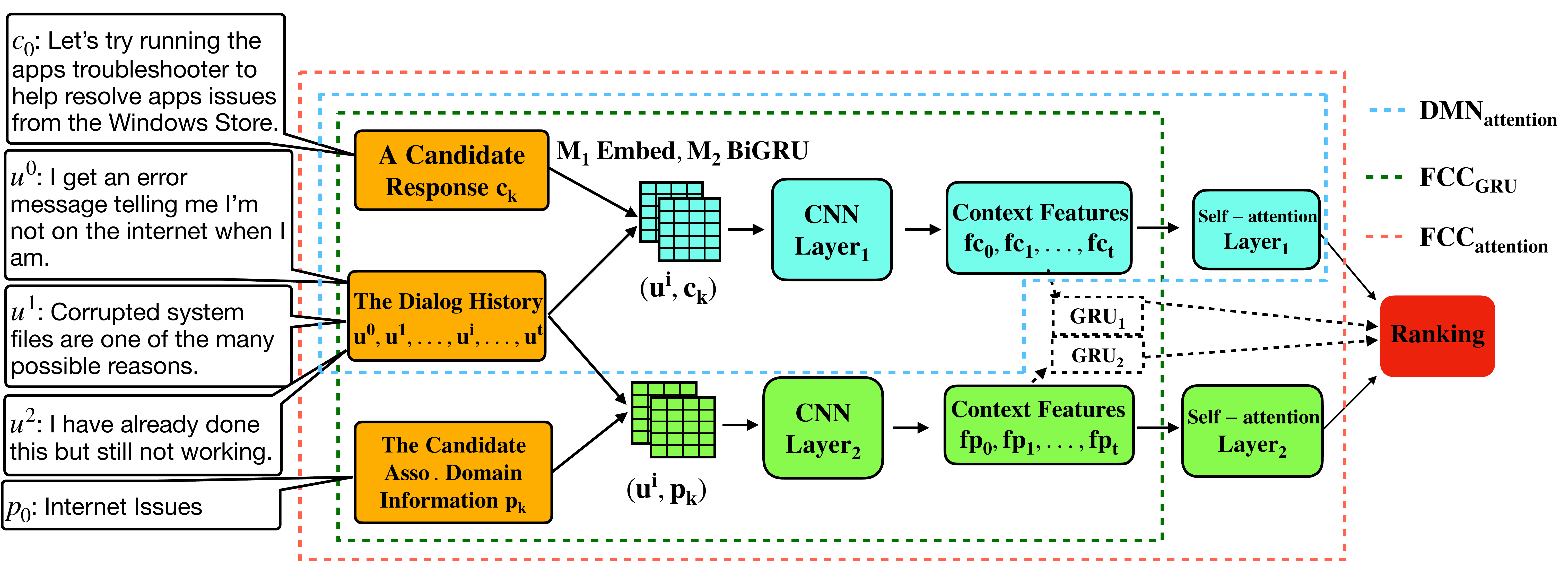}
  \caption{The architectures of the response ranking with domain information and GRU layers ($FCC_{GRU}$), and with domain information and attention layers ($FCC_{attention}$). 
  Symbols denote as follows. $c_0$: the $0^{th}$ candidate the current utterance; $u^i$: the $i^{th}$ utterance in the dialogue, $i\in [0, 1, 2]$; $p_0$: the domain information associated to candidate $c_0$.}
  \label{fig:model-architecture}
\end{figure*}

\section{Approach and Implementation}
\label{sec:approach}

In this section, we, in sequence, define the problem setting (Section~\ref{ssec:problem-formulation}), give an overview of the proposed approach (Section~\ref{ssec:framework-overview}), explain the framework architecture in detail, and present two specific settings for ablation studies (Section~\ref{ssec:model-architecture}). 

\subsection{Task Formulation}
\label{ssec:problem-formulation}
We now formulate the response ranking more formally.
Given a dialogue \textit{D}, at turn \textbf{$t$}, there is a conversation history \{$u^0$, $u^1$,...,$u^t$\}, and a given set of response candidates \{\textbf{$c_0$}, \textbf{$c_1$}, ..., \textbf{$c_j$}, ..., \textbf{$c_k$}\} with their associated domain information \{\textbf{$p_0$}, \textbf{$p_1$}, ..., \textbf{$p_j$}, ..., \textbf{$p_k$}\}, from which they are curated. The instantiated task is to leverage conversation history and candidates' domain information to make ranking decisions to user utterances.

\subsection{Approach Overview}
\label{ssec:framework-overview}

We approach the conversational response ranking problem with a bi-channel end-to-end pipeline, to fuse contextual information both from conversation history and candidate responses. 
First, conversation history interacts with candidate responses and their domain information turn by turn, respectively, to build up interaction representations in each channel. And then, self-attention is applied to each channel to model conversation dependencies. Finally, the output from both channels are concatenated for ranking.

\subsection{Model Architecture}
\label{ssec:model-architecture}
This section first introduces representation modules of the framework, including interaction matrix representation, textual feature representation, and latent ranking representation.
We then describe the specific implementation integrating domain information from candidate provenance besides conversation history, taking advantage of both contextual information sources. Following that, we give the description of self-attention layers on conversation history. Furthermore, we designate two other framework settings for ablation studies.
Finally, we elaborate on the generalization of our model as a flexible framework.

The initial interactions between the two channels adopt the basic structure of the $DMN$ model by \cite{yang2018response} for the following reasons. 
First, the interaction matrix in $DMN$ has its advantage over other text-matching representations \cite{pang2016text}. 
Second, this representation consists of both embedding and hidden state features, which has performed well in the previous state of the art ranking models \cite{yang2020iart}. 
Third, the use of CNN to capture high-level n-gram textual features has been proven to be effective. 
Last, the GRU module can model sequential relationships. Our proposed framework with ablation studies has improvement over the $DMN$ models and is a fair comparison to their performance.

\textbf{\textit{Interaction matrix representation.}} At conversation turn $j$, $u^j$, $c_k$, or $p_k$ is represented by a sequence of word embeddings $E^j_u$, $E^k_c$ or $E^k_p$, and fed into a shared BiGRU to get hidden states, $H^j_u$, $H^k_c$, and $H^k_p$ respectively. 
The embedding interaction matrix between an utterance and a candidate response is calculated by $M_{u_e}^{c_e} = E^j_u \cdot (E^k_c)^T$.
The hidden state interaction matrix is calculated by $M_{u_h}^{c_h} = H^j_u \cdot (H^k_c)^T$.
The same procedure is actualized to have $M_{u_e}^{p_e}$ and $M_{u_h}^{p_h}$ between an utterance and domain information of a candidate response. 

\textbf{\textit{Textual feature representation.}} The interaction matrix representation is fed into a CNN layer, obtaining $c_j^{u,*}$ (* denotes either candidate response $c$ or topic information $p$), the n-gram textual feature representation for each turn in the conversation. 

\textbf{\textit{Latent conversation history representation.}} We have a GRU or a self-attention layer for modeling conversation history.
However, the self-attention layer is more potent in various tasks \cite{vaswani2017attention}. 
This module $DMN_{attention}$ is applied to the comprehensive conversation history features $C_u^*$ = [$c_0^{u,*}$, $c_1^{u,*}$, ..., $c_t^{u,*}$]. 
The hidden states $R_u^*$ = [$r_0^{u,*}$, $r_1^{u,*}$, ..., $r_t^{u,*}$] from the module are concatenated for ranking.

\textbf{\textit{Model architectures.}} Here, we fit the representation modules in the $DMN_{attention}$ model setting, the proposed framework with domain information and GRU layers $FCC_{GRU}$, and that with domain information and attention layers ($FCC_{attention}$). 
All the models are shown in Fig. \ref{fig:model-architecture} with different legends. 
\begin{itemize}
\item The $DMN_{attention}$ model is developed to explore how the self-attention layer affects the ranking performance. 
This model takes candidate responses and dialogue history as input to obtain interaction matrices $M_{u_e}^{c_e}$ and $M_{u_h}^{c_h}$.
The CNN layer takes in interaction matrices and outputs a textual feature representation $C_u^c$.
A self-attention layer is applied to $C_u^c$ to acquire a latent conversation history representation.

\item The $FCC_{GRU}$ model is developed to explore how domain information affects ranking performance. 
It takes candidate responses, their corresponding domain information, and dialogue history to create interaction matrices $M_{u_e}^{c_e}$, $M_{u_h}^{c_h}$, $M_{u_e}^{p_e}$, and $M_{u_h}^{p_h}$.
The CNN layers take in interaction matrices and output textual feature representations $C_u^c$ and $C_u^p$. 
The GRU layers take textual feature representations and output latent conversation history representation $R_u^c$ and $R_u^p$.

\item The $FCC_{attention}$, model follows the same flow as the $FCC_{GRU}$ model, but instead of GRU layers, two self-attention layers are applied to obtain latent conversation history representations.

\item The ranking layer takes in $R_u^c$ for the $DMN_{attention}$ model, and $concat(R_u^c,R_u^p)$ for the $FCC_{GRU}$ and the $FCC_{attention}$ models, and outputs a ranking score for each candidate response.
\end{itemize}

\textbf{\textit{Framework Generalization.}} The domain information and previous utterances can be replaced with, and the parallel structure of the framework can further be expanded to channel in, other contextual features, such as outsourced external knowledge, as an integral part of the end-to-end neural ranking pipeline, to enhance the contextual enrichment.

\textbf{\textit{Framework Summary.}} 
In summary, we presented our new framework for conversational response ranking, \BNAME, which introduces the following new ideas compared to prior work: 1. an introduction of candidate provenance as a new channel to add to conversation history, generating a compact yet comprehensive representation of a dialogue; 2. an implementation of self-attention layers to improve the modeling of multi-turn dependency; 3. our channelized framework easily being expanded to integrate other contextual features in parallel to further enhance contextual enrichment.

\section{Experiments}
\label{sec:experiments}
In this section, we describe experiments in three parts. 
First, we describe the benchmark MSDialog dataset in Section \ref{ssec:dataset}. 
Next, we describe experimental procedures in Section \ref{ssec:experimental-setup}, which include three experiments: 
1. A study on the performance of $FCC_{attention}$;
2. An ablation study comparing a self-attention layer and a GRU layer to model multi-turn dependency;
3. An ablation study on the effect of domain information of candidates on the ranking performance.
Last, we summarize experimental results comparing with the state-of-the-art baselines in Section \ref{ssec:model-evaluation}.

\begin{table*}[!htbp]
\centering
\caption{Comparison of different models over MSDialog. Numbers in bold font mean the result is better compared with the best baseline $IART$ models. $*$ means statistically significant difference over the best baseline $IART_{Bilinear}$ with p $<$ 0.05 measured by the Student’s t-test. $\dag$ means statistically significant difference over $FCC_{GRU}$ model with p $<$ 0.05 measured by the Student’s t-test. $\S$ means statistically significant difference over $DMN$-$PRF$ with p $<$ 0.05 measured by the Student’s t-test.}
\label{tab:result-table}
\begin{adjustbox}{max width=\textwidth}
\begin{tabular}{|l|c|c|c|c|}
\hline
Data          & \multicolumn{4}{c|}{MSDialog}     \\ \hline
Metrics       & R10@1  & R10@2  & R10@5  & MAP    \\ \hline
DMN-KD \cite{yang2018response}        & 0.4908 & 0.7089 & 0.9304 & 0.6728 \\
DMN-PRF \cite{yang2018response}       & 0.5021 & 0.7122 & 0.9356 & 0.6792 \\ \hline
DAM \cite{zhou2018multi}           & 0.7012 & 0.8527 & 0.9715 & 0.8150 \\ \hline
$IART_{Dot}$ \cite{yang2020iart} & 0.7234 & 0.8650 & 0.9772 & 0.8300 \\
$IART_{Outerproduct}$ \cite{yang2020iart}         & 0.7212 & 0.8664 & 0.9749 & 0.8289 \\
$IART_{Bilinear}$ \cite{yang2020iart} & 0.7317 & 0.8752 & 0.9792 & 0.8364 \\ \hline
$DMN_{attention}$ & 0.5544\textsuperscript{\S} & 0.7579\textsuperscript{\S} & 0.9507\textsuperscript{\S} & 0.7180\textsuperscript{\S} \\
\textbf{$FCC_{GRU}$} (our framework)      & \textbf{0.770*} & \textbf{0.8780} & 0.9717 & \textbf{0.8548*} \\
\textbf{$FCC_{attention}$} \ (our framework)& \textbf{0.7879*\textsuperscript{\dag}} & \textbf{0.8992*\textsuperscript{\dag}} & \textbf{0.9810\textsuperscript{\dag}} & \textbf{0.8697*\textsuperscript{\dag}} \\\hline
\end{tabular}
\end{adjustbox}
\vspace{-2.5ex}
\end{table*}

\subsection{Dataset}
\label{ssec:dataset}

The MSDialog conversational dataset is collected from the Microsoft products online forum, which discusses issues in a miscellaneous assortment of domains.  
It includes more than 35,000 conversations and more than 337,000 utterances. 
We use the subset MSDialog-ResponseRank dataset processed by \cite{Qu_2018}. 
In the MSDialogue dataset, candidate responses are extracted from conversations discussing various issues. 
These issues are summarised in the "title" fields in the dataset, which is a fair comparison to domain information of specific components in retrieval-based dialogue systems. 
Therefore, we take "title" fields as our domain information for candidates and this information is reasonably straightforward and easy to get in a dialogue system.

We use Matchzoo\footnote{https://github.com/NTMC-Community/MatchZoo} as the data preprocessing tool. 
Each ranking list, which has one true response and nine candidate responses, is converted to a pair-wise ranking setting.
Each true response will be ranked against each candidate response. 

\begin{figure}[!htbp] 
\centering
  \includegraphics[scale=0.35]{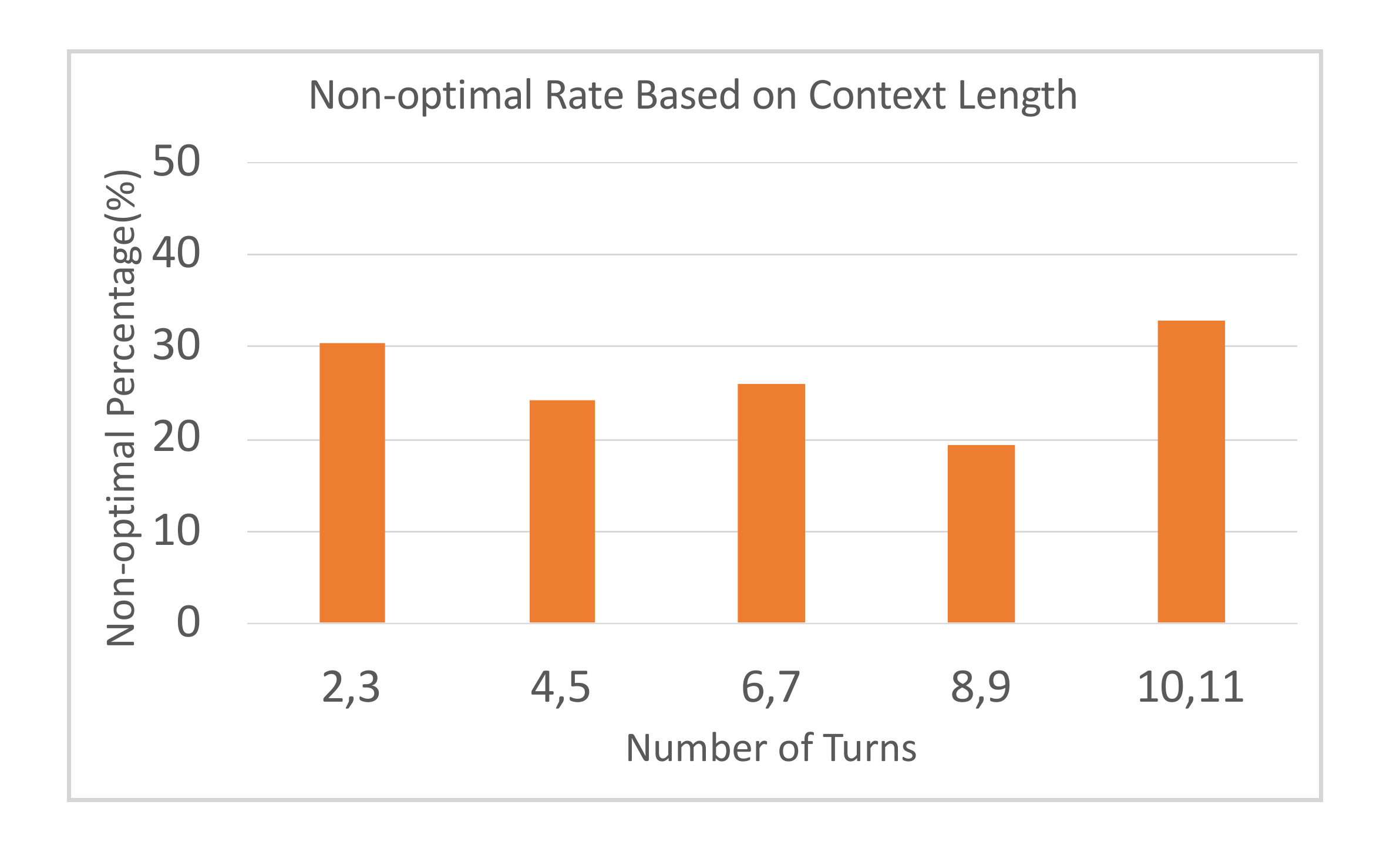}
  \caption{The Non-optimal Rate over Conversation History Length}
  \label{fig:non-optimal-rate}
\end{figure}

\subsection{Experimental Setup}
\label{ssec:experimental-setup}
We have over 173k samples in the training set, and 37k and 35k in the validation and testing sets.
We implement our models using Pytorch \footnote{https://pytorch.org/}. 
For CNN layers, we use two convolution and max-pooling sub-layers with the number of filters [16, 16], convolutional kernels [3, 3], max-pooling kernels [2, 2] and strides [1, 1]. 
The self-attention layer has two heads and two encoder blocks. 
We train on the ranking corpus to gain pre-trained embeddings with dimension 200 with the Word2Vec tool \cite{mikolov2013distributed}.
The maximum number of turns in a dialogue is 10. 
The maximum sequence length for utterance and candidate response is 90 and 30 for the domain information sequence. 
The batch size is 50. We tune all parameters by the validation dataset.

\subsection{Model evaluation}
\label{ssec:model-evaluation}
In this section, we first report the performance of $FCC_{attention}$, comparing with the state-of-the-art baselines in response ranking and response selection fields.
And then, we show the results of ablation studies on the impact of domain information and self-attention layers. 
Experiment results are reported in Table \ref{tab:result-table}.

\textbf{\textit{Main results.}} We evaluate $FCC_{attention}$, on $R10@1$, $R10@2$, $R10@5$, and $MAP$. 
The results show that $FCC_{attention}$, has an improvement on all four metrics over the state of the art $IART$ models, especially on R10@1, R10@2, and MAP, which all have significance p-value $<$ 0.05. 
The performance on recall@1 has the most significant 7.7\% improvement, which is most important since a dialogue system usually picks the best candidate to return to a user. 

\textbf{\textit{The ablation study on domain information.}} To study the impact of domain information compared with the $DMN$ models, we evaluate $FCC_{GRU}$ on the same metrics. 
The results show that with an extra channel to integrate domain information from candidates to the $DMN$ architecture, the ranking performance improves significantly, with margins between 2.2\% to 38.9\% corresponding to different metrics. 
The ranking performance not only surpasses the $DMN$ models but has significant improvements on recall@1 and MAP over $IART$ models, with margins of 5.0\% and 2.2\%.
This comparison confirms the positive effect of domain information from the candidates. The domain information provides 


\textbf{\textit{The ablation study on self-attention layers.}} We conduct an ablation study on the effect of self-attention layers over conversation history. 
The $DMN_{attention}$ model has improvement over the DMN-PRF model with margins ranging from 1.6\% to 10.4\%.
The $FCC_{attention}$ model surpasses the performance of the $FCC_{GRU}$ model with improvement ranging from 1.0\% to 2.4\%.
From the results, it is clear that the self-attention layer impacts positively on the ranking performance.

Furthermore, we analyze the non-optimal rate (percentage of cases in which the true response is not ranked first) as shown in Fig. \ref{fig:non-optimal-rate}, to explore the effectiveness of the self-attention layer conditioning on conversation history length. It is demonstrated that the non-optimal rate drops from about 30\% to 20\% as the length of conversation history increases until a sudden surge on conversations with 10 and 11 (maximum length) turns. 
It is reasonable to conjecture that when the conversation only has a few turns, such as 2 or 3, the model is not fed with enough contextual information to make an optimal decision. While in the opposite case, the model isn't sophisticated enough to isolate effective information from over-long conversations. The self-attention layers are most effective on conversations with 4 to 9 turns.
\section{Discussion and Conclusion}
\label{sec:conclusion}

In this paper, we proposed a flexible framework (\BNAME) capable of incorporating miscellaneous contextual resources for response ranking in multi-turn dialogue systems.
To validate the framework, we implemented embedding domain information of candidates with self-attention layers to improve the relevance modeling between utterances and candidate responses.

Specifically, the domain information adds a second source to interact with utterances, a mechanism to either confirm or alleviate the semantic matching just between conversation history and candidates. One of the examples as a demonstration here:

--Utterance: ...message telling me I am not on the internet while I am ...

--Candidate 1: You will be ...running a trouble shooter... to fix some common issues with Window Update. 
(Domain Info: Adobe Flash Player in Edge and IE is not updating from vulnerability.)

--Candidate 2: Let's try running ... trouble shooter to help resolve app issues from the Windows Store.
(Domain Info: Internet Issues.)

The trained $DMN_{attention}$ model ranked Candidate 1 first, without knowing the domain information. However, FCC models successfully ranked Candidate 2 first since the domain knowledge directly points to the intention of the user. This example clearly supports our claim that domain knowledge from the source of candidates enhances the effectiveness of a response ranking model.

Our overall result supports our claim as well, by outperforming existing state-of-the-art models, with ablation studies to show that both domain information of candidates and self-attention layers lead to critical increments in the performance respectively and conjunctively.

In the future, we would like to investigate on a diversified stream of contextual information feeding into and expanding our framework, and develop hierarchical semantic representations for multi-turn conversations to enrich the information input to improve the capability of modeling longer conversation history.

\bibliographystyle{style/spmpsci}
\bibliography{workshop}
\end{document}